\documentclass[sn-mathphys,Numbered]{sn-jnl}

\usepackage{graphicx}%
\usepackage{multirow}%
\usepackage{amsmath,amssymb,amsfonts}%
\usepackage{amsthm}%
\usepackage{mathrsfs}%
\usepackage[title]{appendix}%
\usepackage{xcolor}%
\usepackage{textcomp}%
\usepackage{manyfoot}%
\usepackage{booktabs}%
\usepackage{algorithm}%
\usepackage{algorithmicx}%
\usepackage{algpseudocode}%
\usepackage{listings}%

\theoremstyle{thmstyleone}%
\theoremstyle{thmstyletwo}%

\theoremstyle{thmstylethree}%

\begin{document}

\title[Article Title]{Guided Scale Space Radon Transform for linear structures detection}


\author*[1]{\fnm{Aicha Baya} \sur{Goumeidane}}\email{a.goumeidane@crti.dz, ab\_goumeidane@yahoo.fr}

\author[2]{\fnm{Djemel} \sur{Ziou}}

\author[1]{\fnm{Nafaa} \sur{Nacereddine}}

\affil[1]{\orgdiv{DTSI}, \orgname{Research Center in industrial Technologies, CRTI }, \orgaddress{\street{P.O.BOX 64}, \city{Algiers}, \postcode{16000},  \country{Algeria}}}

\affil[2]{\orgdiv{D\'epartement informatique}, \orgname{Universit\'e de Sherbrooke }, \orgaddress{\city{ Sherbrooke}, \state{Qu\'ebec},  \country{Canada}}}


\abstract{Using integral transforms to the end of lines detection in images with complex background, makes the detection a hard task needing additional processing to manage the detection. As an integral transform, the Scale Space Radon Transform (SSRT) suffers from such drawbacks, even with its great abilities for thick lines detection. In this work, we propose a method to address this issue for automatic detection of thick linear structures in gray scale and binary images using the SSRT, whatever the image background content. This method involves the calculated Hessian orientations of the investigated image while computing its SSRT, in such a way that linear structures are emphasized in the SSRT space. As a consequence, the subsequent maxima detection in the SSRT space is done on a modified transform space freed from unwanted parts and, consequently, from irrelevant peaks that usually drown the peaks representing lines. Besides, highlighting the linear structure in the SSRT space permitting, thus, to efficiently detect lines of different thickness in synthetic and real images, the experiments show also the method robustness against noise and complex background.}




\maketitle

\section{Introduction}\label{sec1}

Automatic linear structures detection in digital images is a classical topic addressed by many researchers for years. Each research has tackled the issue from a different point of view, exploiting different techniques leading to application-based approaches. Indeed, images containing linear structures can be produced with various image formation process, and therefore, can be very different from each other?s, having, consequently, different characteristics, so the proposed approach must, therefore, manage the image content. 
As previously noticed, automatic detection of lines has been employed in a wide range of applications, including robot guidance \cite{chen2023,liu2022}, Autonomous vehicles and Advanced Driver Assistance Systems \cite{zak2023,fan2019,luo2020}, high-voltage lines detection \cite{luo2021}, Barcode detection \cite{hansen2017}, road network extraction \cite{hikosaka2022,xu2022}. The exploited techniques vary from deep Learning to integral transforms-driven methods or their association \cite{zhao2021}, as well as, derivatives-based proposals \cite{li2023}. However, the most robust approaches to achieve the task of straight line detection are the ones involving integral transforms as Radon Transform (RT)\cite{helgason1999}, Hough Transform (HT) \cite{hough}, and the transforms derived therefrom associated to basic image processing \cite{suder2021}. In fact, RT and HT and their extensions have been widely applied to the above-mentioned purpose. They consist in converting the global line detection problem in the image domain into a peak detection problem in the transform one, even in presence of noise \cite{nacereddine2014}. These peaks positions correspond to the lines parameters. Unfortunately, most of the integral transform-based proposed methods include pre-processing steps, additional tricks, supplementary procedures and predefined thresholds to manage the detection, as often ambiguities regarding the relevant and irrelevant peaks in the transforms spaces rise. In fact, applied to images, these transforms correspond to projections and pixels intensity accumulation over the image in all directions. Consequently, sinograms or accumulators can present peaks where there are no linear features. To overcome such drawbacks, some proposals suggest to limit the computation to dynamically adjusted Region of Interests (ROI) in the image obtained with deep learning, with the aim of avoiding irrelevant image parts \cite{shen2021}. Others propose to exclude a large number of useless peaks in the transform space by keeping only the ones related to lines passing through the vanishing point in lane detection application \cite{zheng2018}, whereas voting schemes is exploited to select the real peaks among the others by using statistics and entropy in \cite{xu2015} and \cite{xu2014}, respectively. Concerning RT and HT, it is worth to note that their respective spaces are constructed by pixels values accumulation (in a line-wise manner) done line by line which make them adequate only for filiform lines, unless involving additional processing and artifices to do the job accuratly for thick structures as proposed by the authors in \cite{zhang2007}. Fortunately, the work proposed in \cite{Ziou} have introduced a novel integral transform called Scale Space Radon transform (SSRT), which can be viewed as a significant generalized form of the Radon Transform by replacing the Dirac function in RT by a Gaussian kernel tuned by a scale space parameter. In the early paper about the SSRT, the authors have shown that this transform can be used to detect accurately and in an elegant manner thick lines when the scale parameter is tuned correctly. Nevertheless, the SSRT has inherited some RT drawbacks, especially its sensitivity to complex backgrounds, particularly when the feature is too short comparing to the image dimensions. This fact produces peaks in the SSRT space that do not correspond to lines. To overcome this shortcoming, authors in \cite{goumeidane2021} have proposed to construct the SSRT space only around pre-computed SSRT parameters yielded with multiscales Hessian of an image, with the aim of discarding orientations and positions bringing useless information. However, the pixels values integration is done inside a band for the SSRT in all directions in the entire image, which increases the interference of the irrelevant parts of the image when constructing the SSRT space, leading, as a result, to the introduction of a bias in the detected linear structures. This situation occurs even if this space is constructed around precomputed  SSRT parameters and excludes the remaining of the space, because still the transform space computation incorporates image elements parts that lay in the precomputed orientation and position of the researched linear structures. These image elements parts will interfere while computing the SSRT space and may deviate the detected line from its correct position, especially when the processed images are gray valued and/or the linear structure is a short one.
In this paper, we propose a solution to this issue to, simultaneously, emphasis the linear features in the SSRT space and exclude major or all parts of the other features. Furthermore, in each SSRT projection computation, the proposed approach removes the contribution of all image elements that do not have the orientation of the actual projection. These procedures operate by the means of the SSRT and the Hessian of the processed image.

The remainder of this paper is presented as follows. In Sect. 2 we introduce the SSRT transform and the image Hessian. Section 3 is consecrated to the proposed method for linear structure exatraction. Section 4 is dedicated to experiments and results. Finally, conclusion is drawn in Sect. 5.

\section{Methods and material}\label{sec2}
\subsection{Scale Space Radon Transform}
The Scale Space Radon Transform (SSRT), $\check{S_f}$, of an image $f$, is a matching of a kernel and an embedded parametric shape in this image. If the parametric shape is a line parametrized by the location parameter $\rho$ and the angle $\theta$ and the kernel is a Gaussian one, then $\check{S_f}$ is given by \cite{Ziou}
\begin{equation}\label{ssrt}
\check{S_f}(\rho,\theta,\sigma)=\frac{1}{\sqrt{2\pi}\sigma} \int_{\cal X} \int_{\cal Y} f(x,y)e^{-\frac{(x\cos \theta +y\sin \theta -\rho)^2}{2\sigma^2}}dxdy
\end{equation}
Here, $\sigma$ is the scale space parameter. It returns that the RT $\check{R_f}$  is a special case of $\check{S_f}$ (when $\sigma\rightarrow \ 0$) \cite{Ziou}. The replacement of the Dirac function $\delta$ in RT by a Gaussian kernel in the SSRT
allows $\check{S_f}$ to handle correctly embedded shapes even if they are not filiform, unlike $\check{R_f}$. The role of the chosen kernel is to control the parametric shape position inside the embedded object via the scale parameter and then, the detection is reduced to maxima detection in the SSRT space. 

\begin{figure}[h]
	\centering
	\includegraphics[width=75mm]{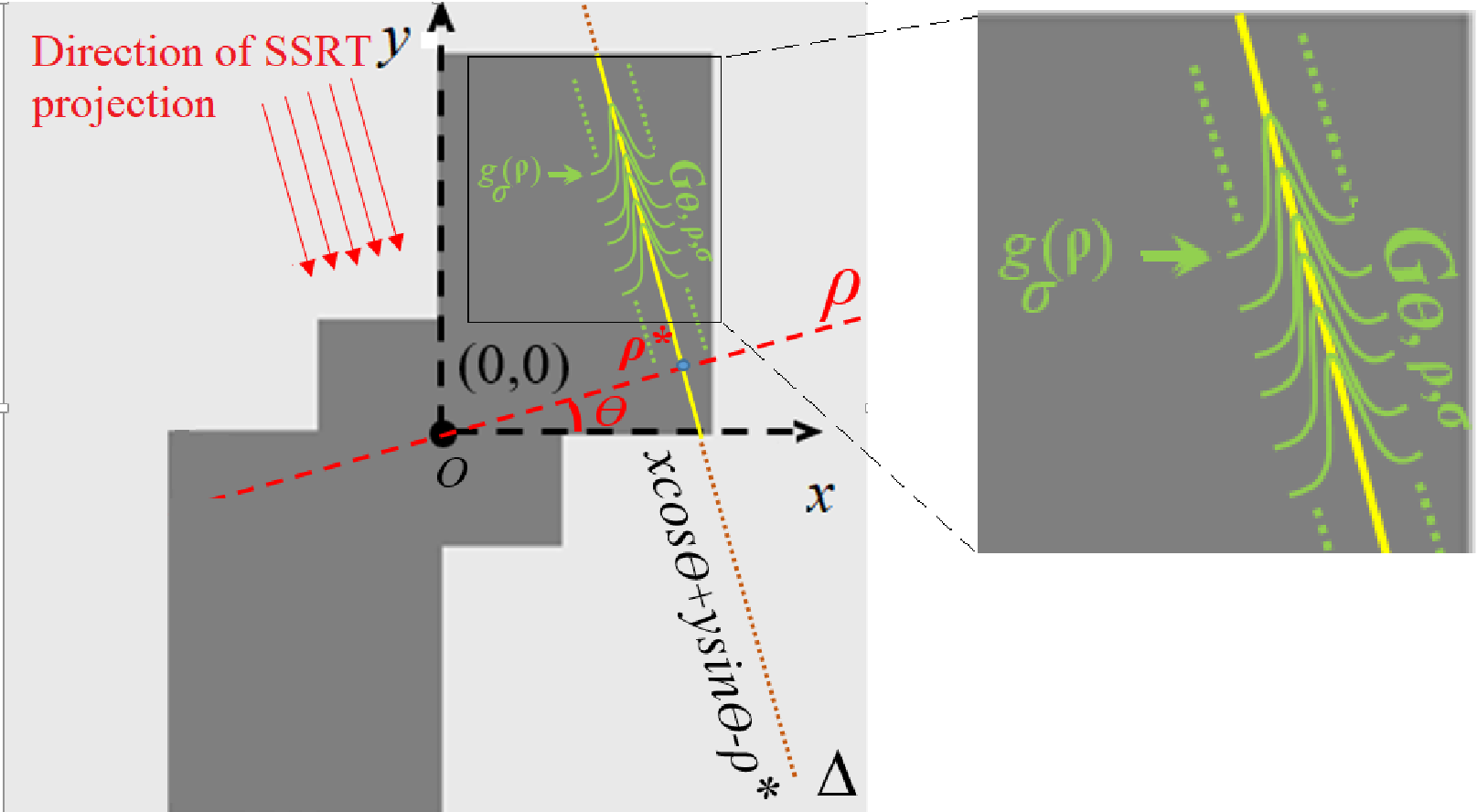}
	\caption{Directional Gaussian $DG_{\theta,\rho,\sigma}$ used in SSRT computation, the unidimensional Gaussian $g_\sigma(\rho)$ and direction of SSRT projection defined with red arrows. }\label{DG}	
\end{figure}
The directional Gaussian in \eqref{ssrt}, $DG_{\theta,\rho,\sigma}(x,y)=\frac{1}{\sqrt{2\pi}\sigma} e^{-\frac{(x\cos\theta+y\sin\theta-\rho)^2}{2\sigma^2}}$, can be viewed in Fig.\ref{DG}. It is an infinity of parallel 1d Gaussians, $g_{\sigma}(\rho)$ having their mean values belonging to the line $\Delta$ with equation $ x\cos\theta +y\sin\theta -\rho =0$.

\begin{figure}[h]
	\centering
	\includegraphics[width=95mm]{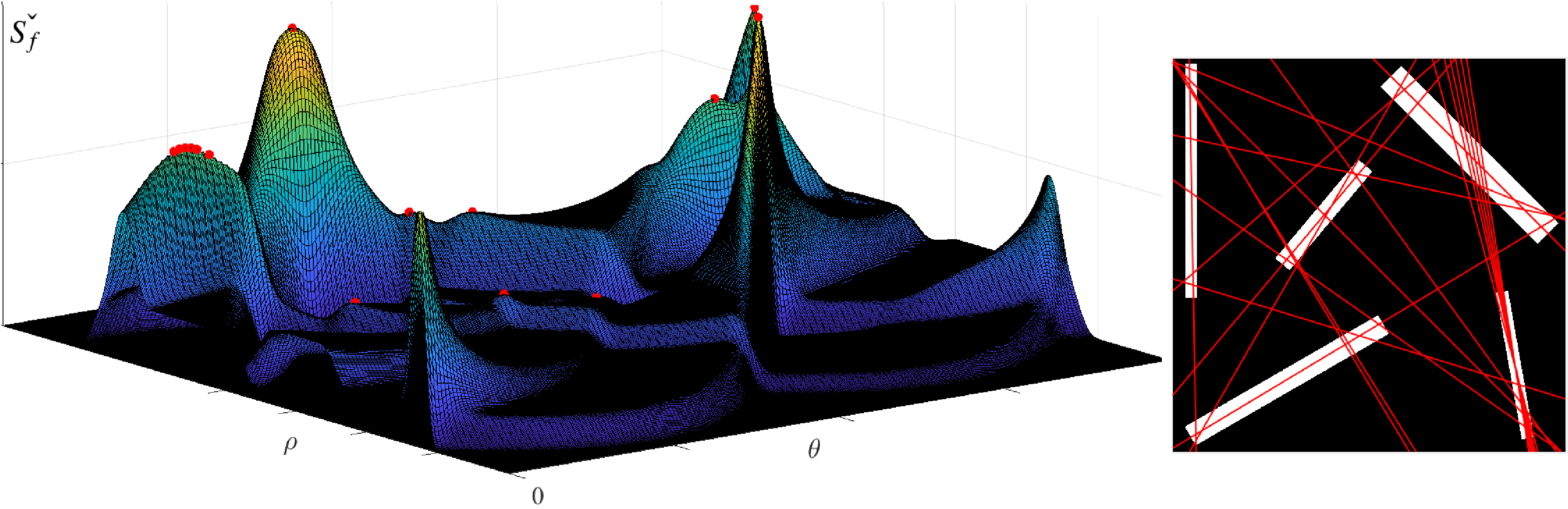}
	\caption{From left to right: SSRT space with highlighted maxima and the corresponding line. }\label{ssrtbars}	
\end{figure}

\begin{figure}[h]
	\centering
	\includegraphics[width=95mm]{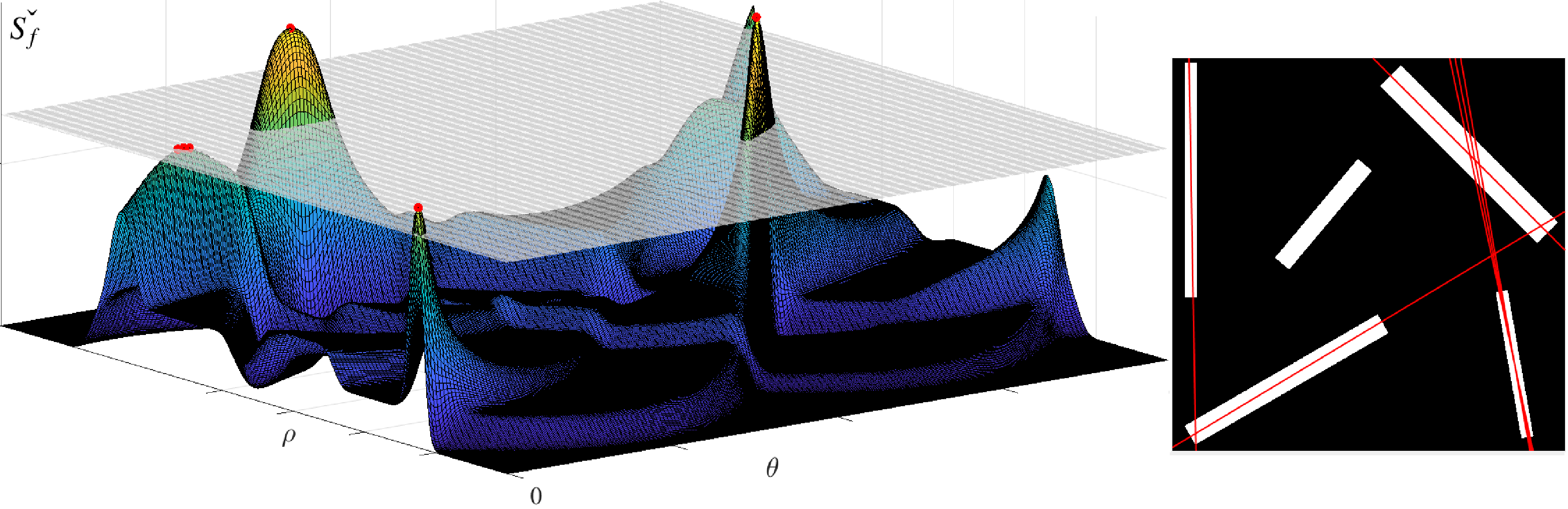}
	\caption{From left to right: Thresolded SSRT space wit the remaining maxima and the corresponding lines }\label{ssrtseuil}	
\end{figure}

Detection of linear structures performed with SSRT employed alone, gives accurate results when the image is not complex and the SSRT space present maxima related to the available linear structures as shown in \citep{Ziou} Nevertheless, as a detection method, it can be subjected to problems related to the detection issues, i.e: over-detection, under-detection and accuracy. The over-detection occurs when the number of linear structures increases or when the background is a complex one. In fact, in such cases, the SSRT detection becomes a complicated task and we can notice, therefore, the apparition of maxima that do not represent structures. As a result, spurious linear structures are detected, as depicted in Fig.\ref{ssrtbars}. Furthermore, when applying a "naive" thresholding to remedy to the over detection and improve the detection, not all spurious lines are deleted, while suppressing relevant ones, as shown in Fig.\ref{ssrtseuil}, this is the under detection.

\begin{figure}[h]
	\centering
	\includegraphics[width=50mm]{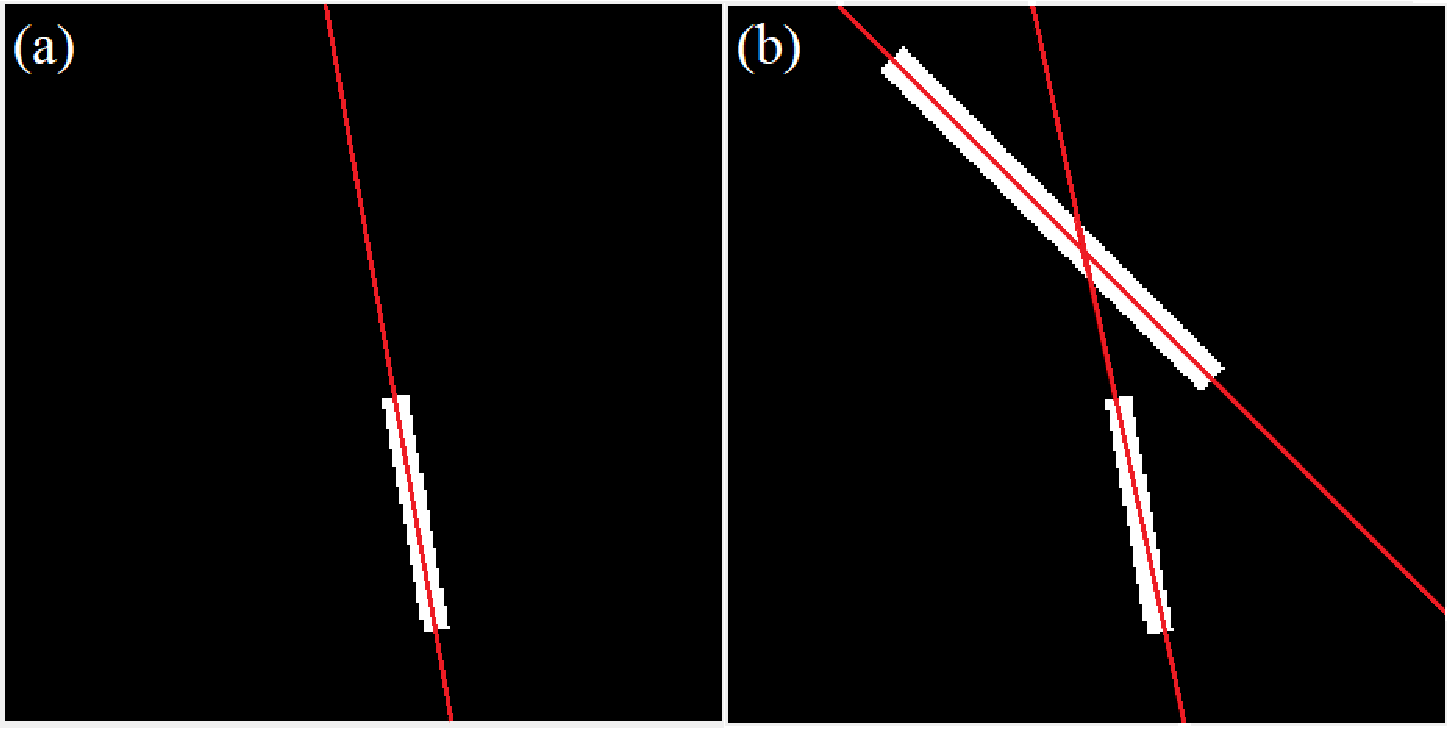}
	\caption{(a): Detection with one structure. (b) Detection when adding in a certain way another structure.}\label{ssrt_detect}	
\end{figure}
\begin{figure}[h]
	\centering
	\includegraphics[width=95mm]{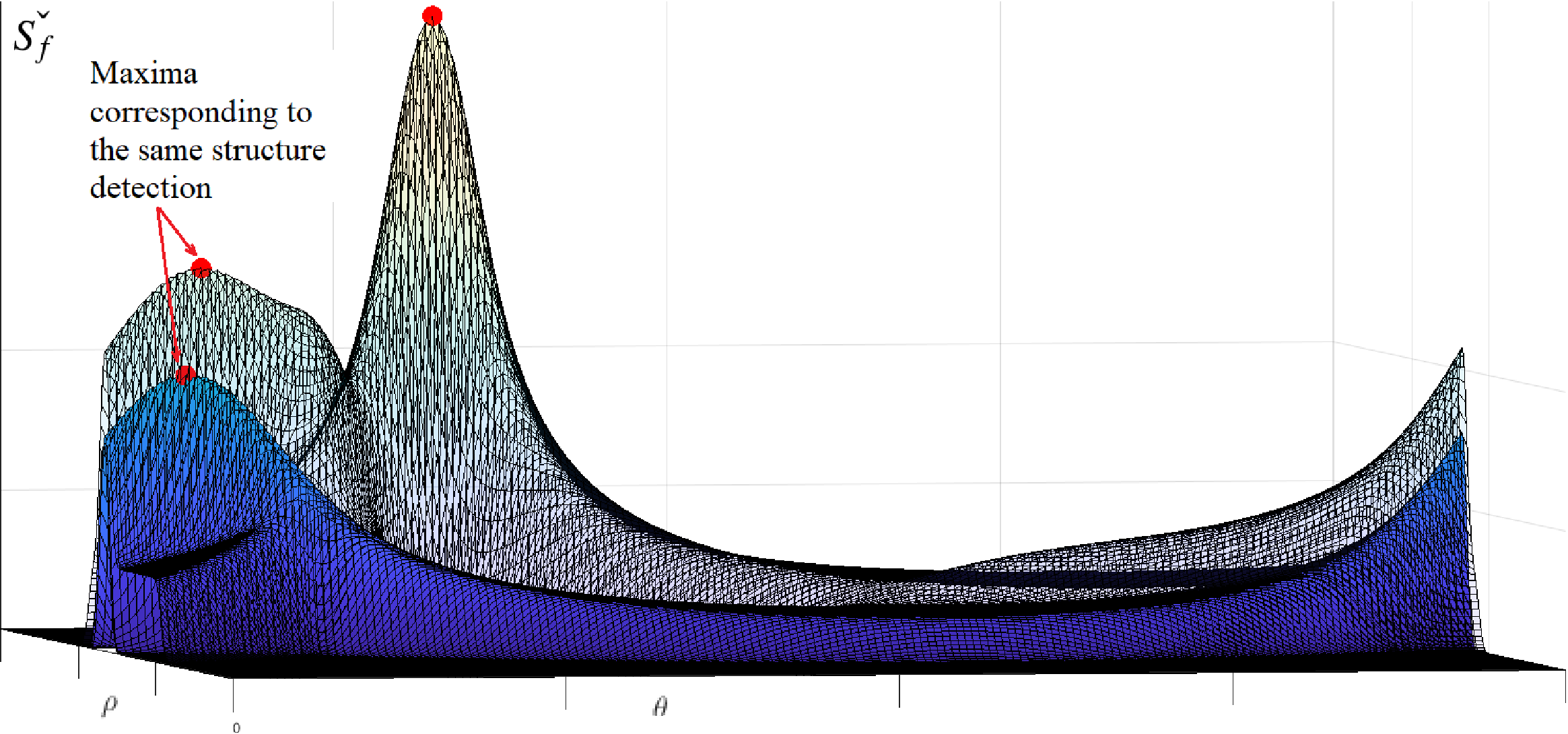}
	\caption{Superposition of the blue SSRT space of the image in Fig.\ref{ssrt_detect}.a and the transparent one of the image in Fig.\ref{ssrt_detect}.b, and highlighted maxima. }\label{2SSRT}	
\end{figure}
\begin{figure}[h]
	\centering
	\includegraphics[width=73mm]{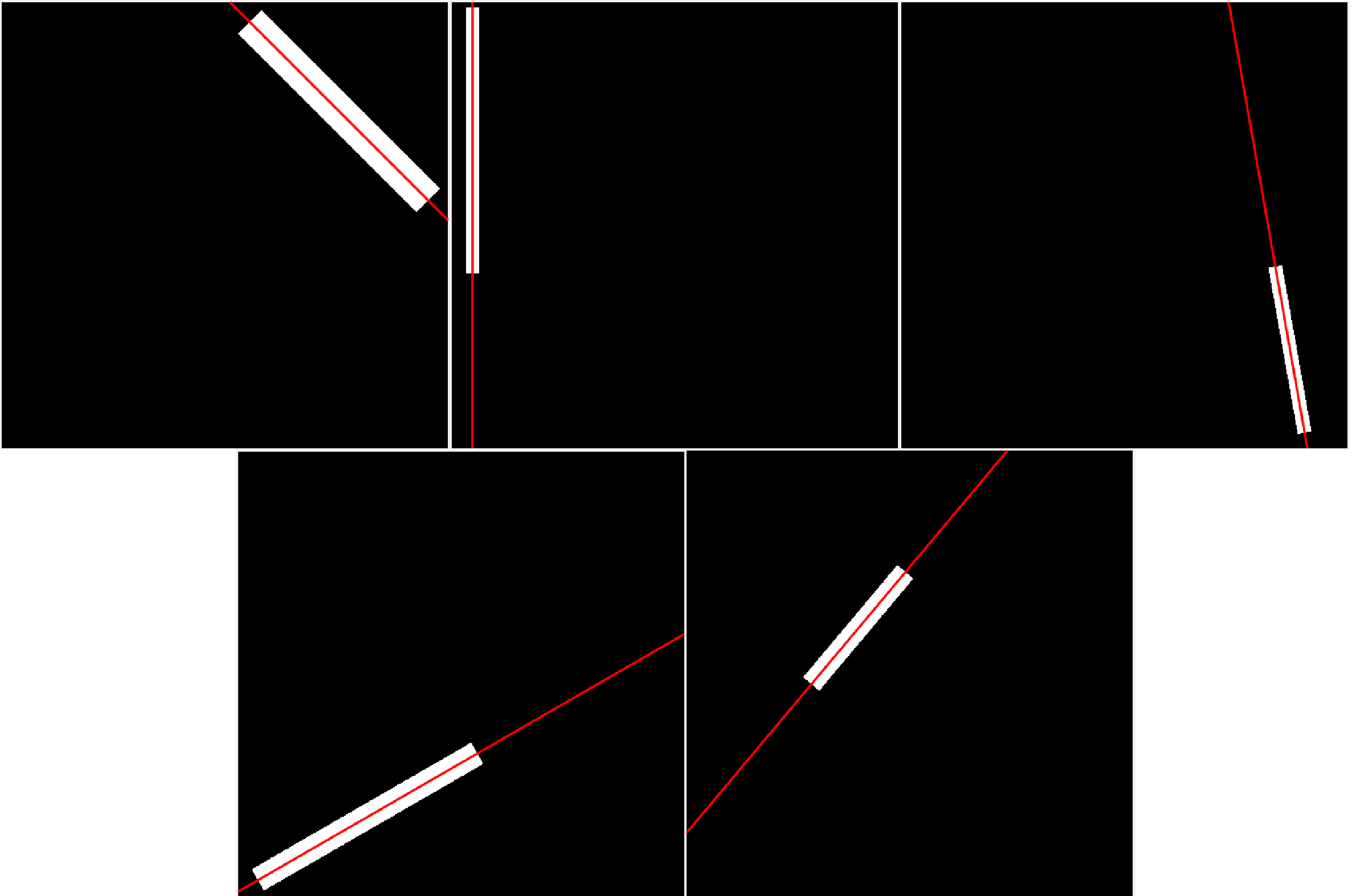}
	\caption{Structures centerlines detections when each structue is alone in the image }\label{5bars}	
\end{figure}

Moreover, some configurations, as shown by the image in Fig.\ref{ssrtbars} and in Fig.\ref{ssrt_detect}.b. lead to the deviation of the detected lines. For example in Fig.\ref{ssrt_detect}.b, the longest linear structure, being in the continuation of the short one with slightly different angle, has displaced the maximum of the latter in the SSRT space and therefore its position in the image domain. We can see in Fig.\ref{ssrt_detect}.a, that the detection is accurate when the short thick line is alone. Regrading the SSRT spaces of the two images, they are superposed and depicted in Fig.\ref{2SSRT} where we can see that the highlighted maximum in the blue SSRT space (corresponding to Fig.\ref{ssrt_detect}.a) has been subjected a modification (the transparent space corresponds to Fig.\ref{ssrt_detect}.b.) when adding the other structure in Fig.\ref{ssrt_detect}.b. This deformation has displaced the short structure maximum, and create a problem of accuracy.
The same thing can be said about the structures in Fig.\ref{ssrtbars}. In fact when each structure is processed alone, the SSRT centerline detection result is the adequate one, as seen in Fig.\ref{5bars}.

Consequently, some additional processing must be added to the SSRT detection to deal with such situations and improve the detection. To this end, we joint the image Hessian computation to the SSRT as we will see later.
\subsection{The Hessian}
The Hessian of an image has been used for identifying particular structures centers when the scale $\sigma_h$ of the Hessian matches the size of the local structures in images ~\citep{Steger}. The Hessian matrix $ H_{\sigma_h}$ of the intensity image $f$ at a scale $\sigma_h$ is given by
\begin{equation}
\small
H_{\sigma_h} = \begin{pmatrix} H_{11}(\sigma_h) & H_{12}(\sigma_h) \\ H_{21}(\sigma_h) & H_{22}(\sigma_h) \\ \end{pmatrix}= \begin{pmatrix}  \sigma_h^2f\circledast \frac{\partial^2 }{\partial x^2}G(\sigma_h) & \sigma_h^2f\circledast \frac{\partial^2 }{\partial x \partial y}G(\sigma_h)\\ \sigma_h^2f\circledast \frac{\partial^2 }{\partial y \partial x}G(\sigma_h) & \sigma_h^2f\circledast \frac{\partial^2 }{\partial y^2}G(\sigma_h) \\ \end{pmatrix} \normalsize
\end{equation}
$G(\sigma_h)$ is the bivariate Gaussian kernel of standard deviation $\sigma_h$ and $\circledast$ is the convolution operator. Under continuity assumptions the Hessian matrix is symmetric and non-negative definite. 

\begin{figure}[h]
	\centering
	\includegraphics[width=100mm]{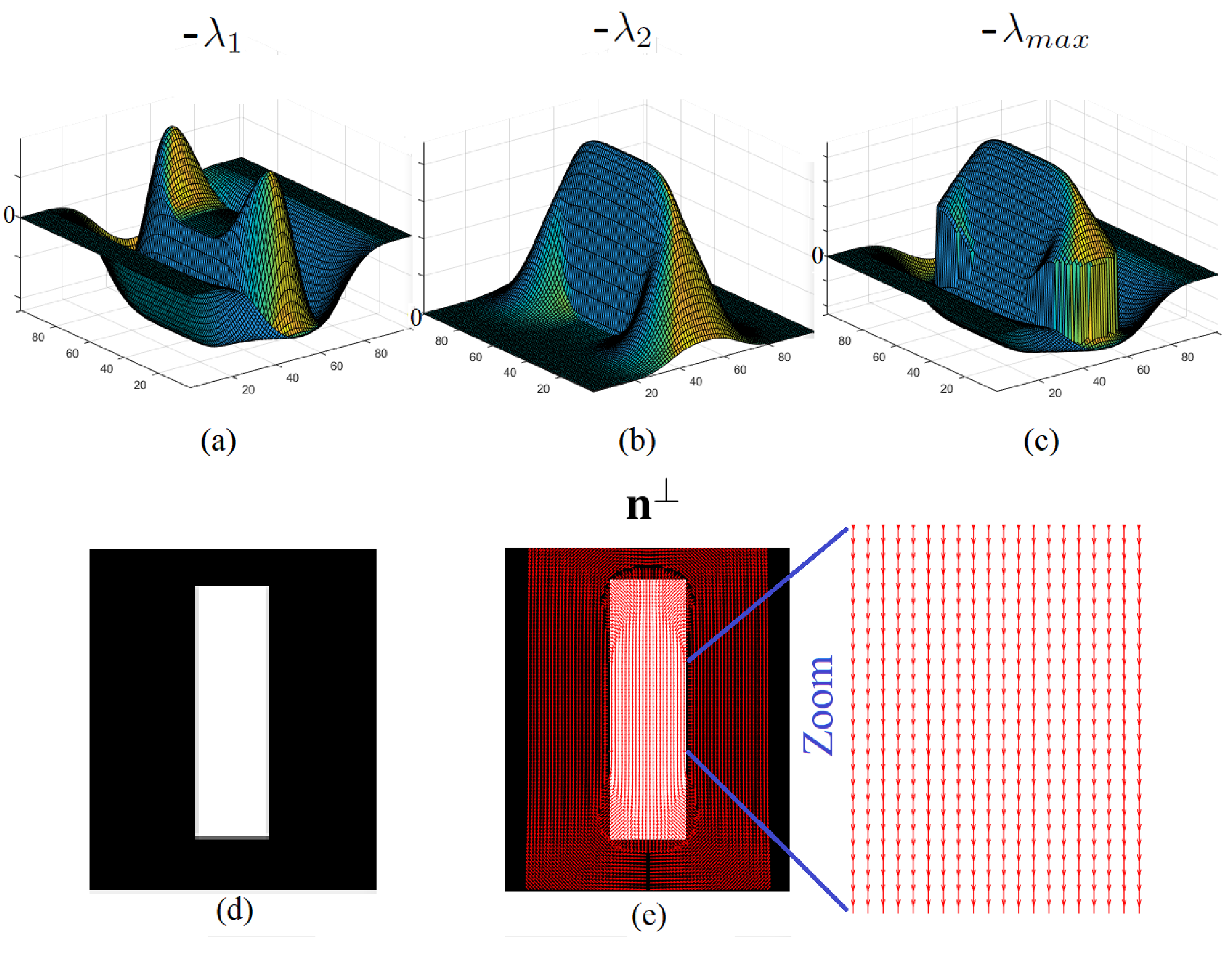}
	\caption{3D representation of Hessian components; (a) $-\lambda_1$, (b) $-\lambda_2$, (c) $-\lambda_{max}$, of the linear structure in (d). (e) Superposition of the linear structure and the vectors field $\textbf{n}^\perp$ with a zoomed part of it  }\label{lambdasn}	
\end{figure}

Let  $\lambda_1$ and $\lambda_2$ be the eigenvalues of $H_{\sigma_h}$ obtained by eigenvalue decomposition as given in ~\citep{deschenes2004}. Let, furthermore, $\lambda_{max}$ be the eigenvalue with maximum absolute value at a point $(x,y)$ and $\textbf{n}(n_x,n_y)$ the unitary vectors field, designating the orientations ${\phi}$ associated with $\lambda_{max}$. The components $n_x$ and $n_y$ are the components of $\textbf{n}$ along the $x$-axis and the $y$-axis respectively. They are computed as  $\small
\left[n_x(x,y),n_y(x,y)\right]=\left[\frac{H_{11}(x,y), \ \lambda_{max}(x,y)-H_{11}(x,y)}{\sqrt{H_{21}(x,y)^2+(\lambda_{max}(x,y)-H_{21}(x,y))^2}}\right] \normalsize$. 
Furthermore, let $\textbf{n}^\perp$ be the vectors field made of vectors normal to those composing $\textbf{n}$. In case of linear structures, a part of the field $\textbf{n}^\perp$ matching the inside portion of the structure, points towards to the orientation of the latter. Before going further, it is worth to mention that here, and for notation, we have chosen that $|\lambda_2|\geq |\lambda_1|$. Consequently, for a bright linear structure on a dark background, we can see a 3D representations of $\lambda_1$, $\lambda_2$ and $\lambda_{max}$ in Fig.\ref{lambdasn}.a, Fig.\ref{lambdasn}.b and Fig.\ref{lambdasn}.c, of the linear structure in Fig.\ref{lambdasn}.d, as well as the field $\textbf{n}^\perp$ and a zoom on it, inside the structure in Fig.\ref{lambdasn}.e .

\begin{figure}[h]
	\centering
	\includegraphics[width=100mm]{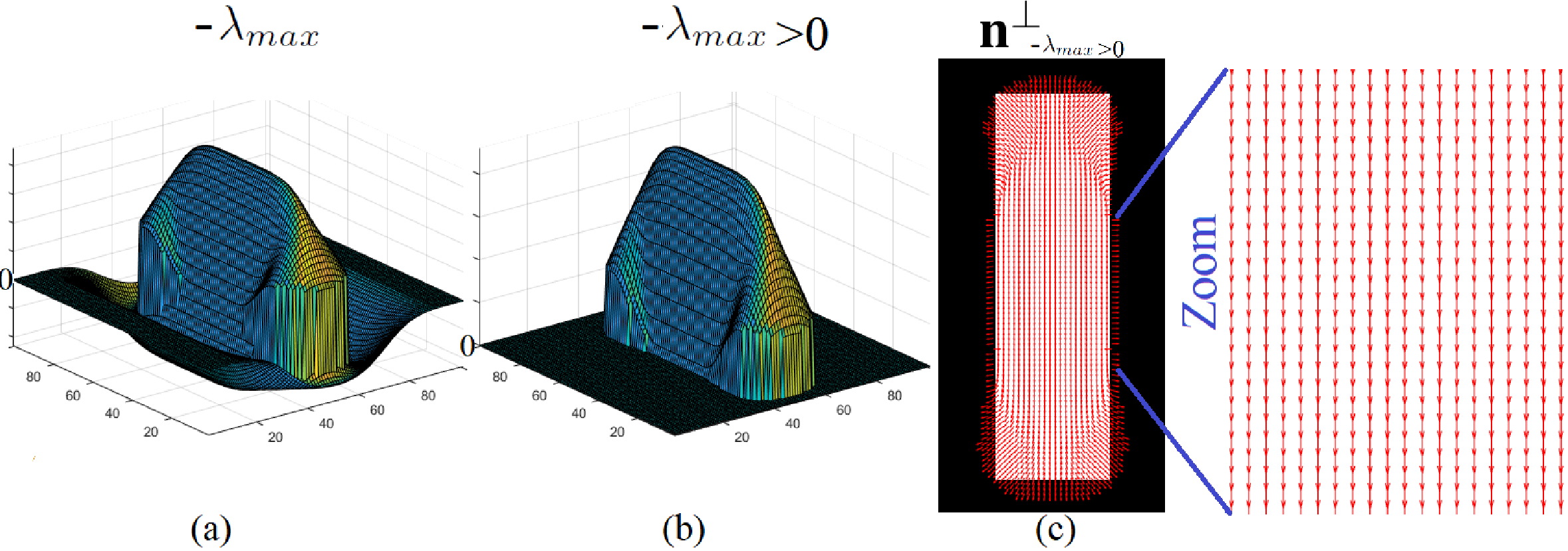}
	\caption{(a):$-\lambda_{max}$, (b):$-\lambda_{max}>0$, (c): the corresponding vectors field $\textbf{n}^\perp_{-\lambda_{max}>0}$ for the linear structure in Fig.\ref{lambdasn}.d. }\label{vectors}	
\end{figure}

If one wants to consider only vectors $n^\perp(x,y)$ inside the structures then, the ones matching the outside of the structure can be discarded by keeping only those corresponding to $-\lambda_{max}(x,y)>0$, i.e ; if $-\lambda_{max}(x,y)<0 $, then $\textbf{n}^\perp(x,y) \leftarrow 0$ as shown in Fig.\ref{vectors}. Let us call the resulting field, illustrated in \ref{vectors}.c, $\textbf{n}^\perp_{-\lambda_{max}>0}$.
\section{Guided SSRT}\label{sec3}
For the sake of simplicity, let us rename, in the following, $\textbf{n}^\perp_{-\lambda_{max}>0}$ as  $\textbf{n}^\perp$.
To take advantage of the SSRT benefits without being subjected to the drawbacks mentioned before, SSRT computation will be guided by the orientation of the linear structures. This orientation is provided via the vector field $\textbf{n}^\perp$ computed with the Hessian of the image under investigation $f$. Let us call this SSRT guided by the Hessian of the image $f$, $\check{S_{f_H}}$. 
We define $\textbf{p}_\theta$, the unitary vector field applied on all image points. The directions of vectors composing $\textbf{p}_\theta$ is the one of the actual SSRT projection as seen in Fig.\ref{ssrt}. If $\Gamma(x,y)=|(\textbf{n}^\perp(x,y)\odot \textbf{p}_\theta(x,y))|$, where $\odot$ is the dot product operator, let us, moreover, define $f_H$, the image yielded by the operation $f_H(x,y)=f(x,y)\Gamma(x,y)$. The function $\Gamma(x,y)$ can be written as $\Gamma(x,y)=|{p_\theta}_x(x,y)n^\perp_x(x,y)+{p_\theta}_y(x,y)n^\perp_y(x,y)|$, where ${p_\theta}_x$, $n^\perp_x$, ${p_\theta}_y$ and $n^\perp_y$ are vectors $\textbf{n}^\perp$, $\textbf{p}_\theta$ components along the $x$-axis and the $y$-axis, which makes this function taking its values in $\left[0\ 1 \right]$. If fact, when the directions of the vectors $\textbf{n}^\perp(x,y)$ and $ \textbf{p}_\theta (x,y)$ are approximately of the same order in absolute value, then $\Gamma(x,y) \approx 1 $, but when their difference approximates in absolute value $\pi/2$, then, $\Gamma(x,y) \approx 0 $. We can see that $f_H$ changes for each projection orientation , as $\textbf{p}_\theta$ depends on the latter. Consequently, in each SSRT projection, all image features in $f$ that have not as orientation the actual projection direction, are forced down in $f_H(x,y)$. Furthermore, to lessen further the influence of previously mentioned image parts, $\Gamma(x,y)$ is raised to a power $M$, with $M>>1$, dragging values of all pixels in $f$ of location (x,y) to zero in $f_H$ when ${p_\theta}_x(x,y)\ne n^\perp_x(x,y)$ and ${p_\theta}_y(x,y)\ne n^\perp_y(x,y)$, enhancing meanwhile, image features having SSRT projection direction as orientation. It worth mentioning here, that vectors in $\textbf{n}^\perp(x,y)$ are normalized so that their respective magnitude is equal to 1, making them unitary vectors.

\begin{figure}[h]
	\centering
	\includegraphics[width=90mm]{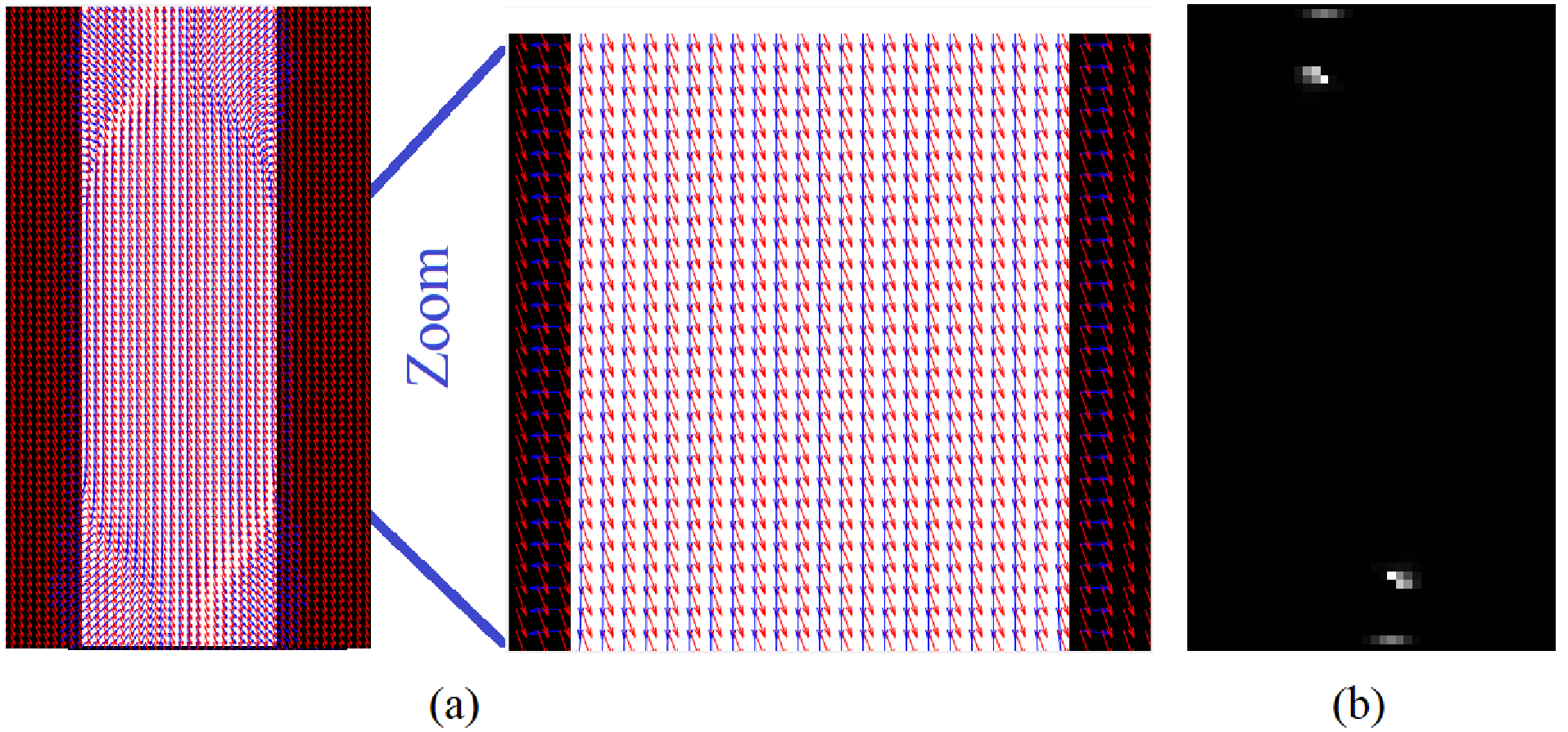}
	\caption{(a) Superposition of the fields $\textbf{n}^\perp$ (in blue) and field $\textbf{p}_\theta$ (in red) for $\theta=20$ of the linear structure depicted in Fig.\ref{lambdasn}.d, and (b) the corresponding image $f_H$ }\label{FH}	
\end{figure}
Fig.\ref{FH} shows both $\textbf{n}^\perp$ and $\textbf{p}_\theta$ for $\theta=20$ degrees as well as the corresponding $f_H$ for $M$ set equals to 100. We can see in this case that the linear structure has almost disappeared in $f_H$, i.e. almost pixels are set to zero (0) , which means that the projection becomes almost blind regarding this image content for this orientation. The SSRT $\check{S_{f_H}}$ exploiting $f_H$ is computed as

\begin{equation}\label{ssrtfH}
\check{S_{f_H}}(\rho,\theta,\sigma)=\frac{1}{\sqrt{2\pi}\sigma} \int_{\cal X} \int_{\cal Y}f_H(x,y)e^{-\frac{(x\cos \theta +y\sin \theta -\rho)^2}{2\sigma^2}}dxdy
\end{equation}

\subsection{Hessian scale}

\begin{figure}[h]
	\centering
	\includegraphics[width=60mm]{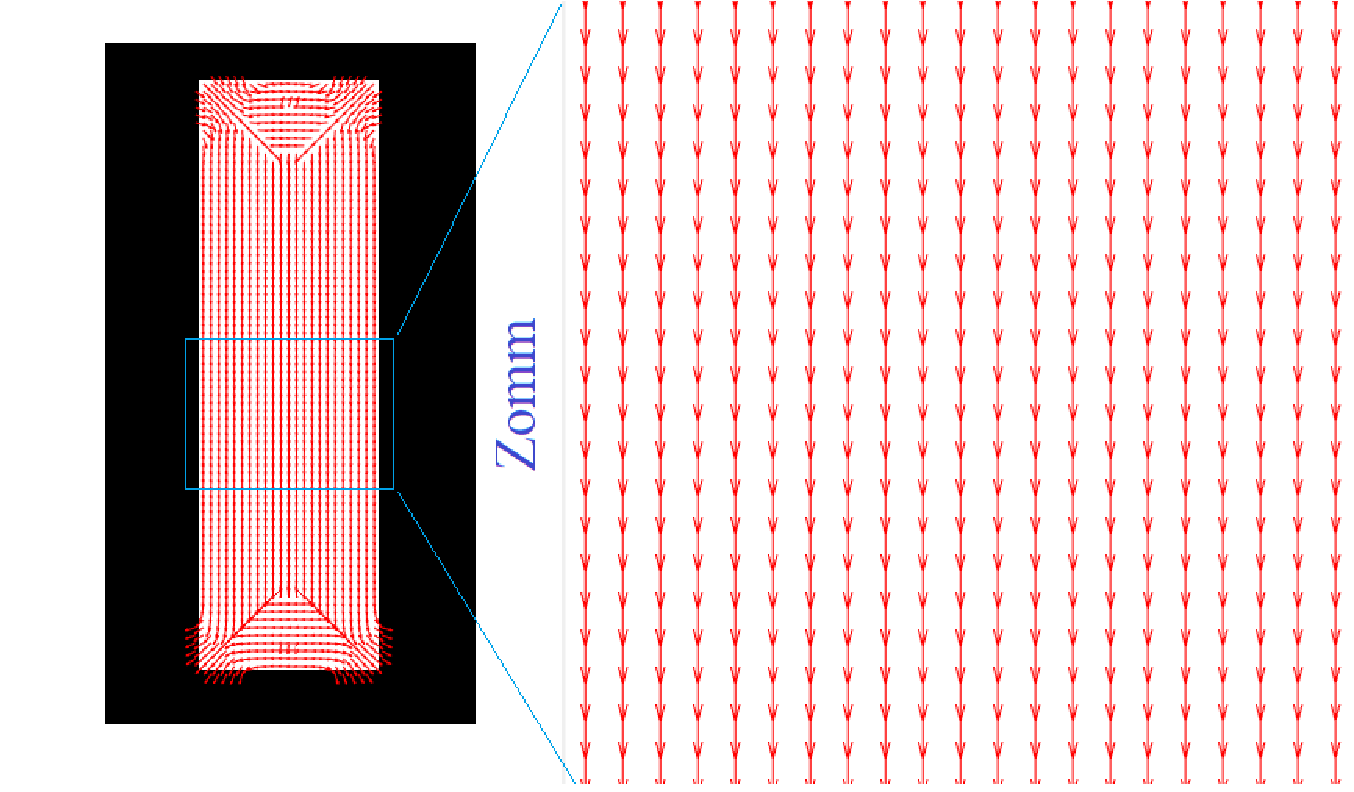}
	\caption{ (a) the field $\textbf{n}^\perp$ generated with $\sigma_h=3.28$ and a zoom on it}\label{Hess1}	
\end{figure}

\begin{figure}[h]
	\centering
	\includegraphics[width=80mm]{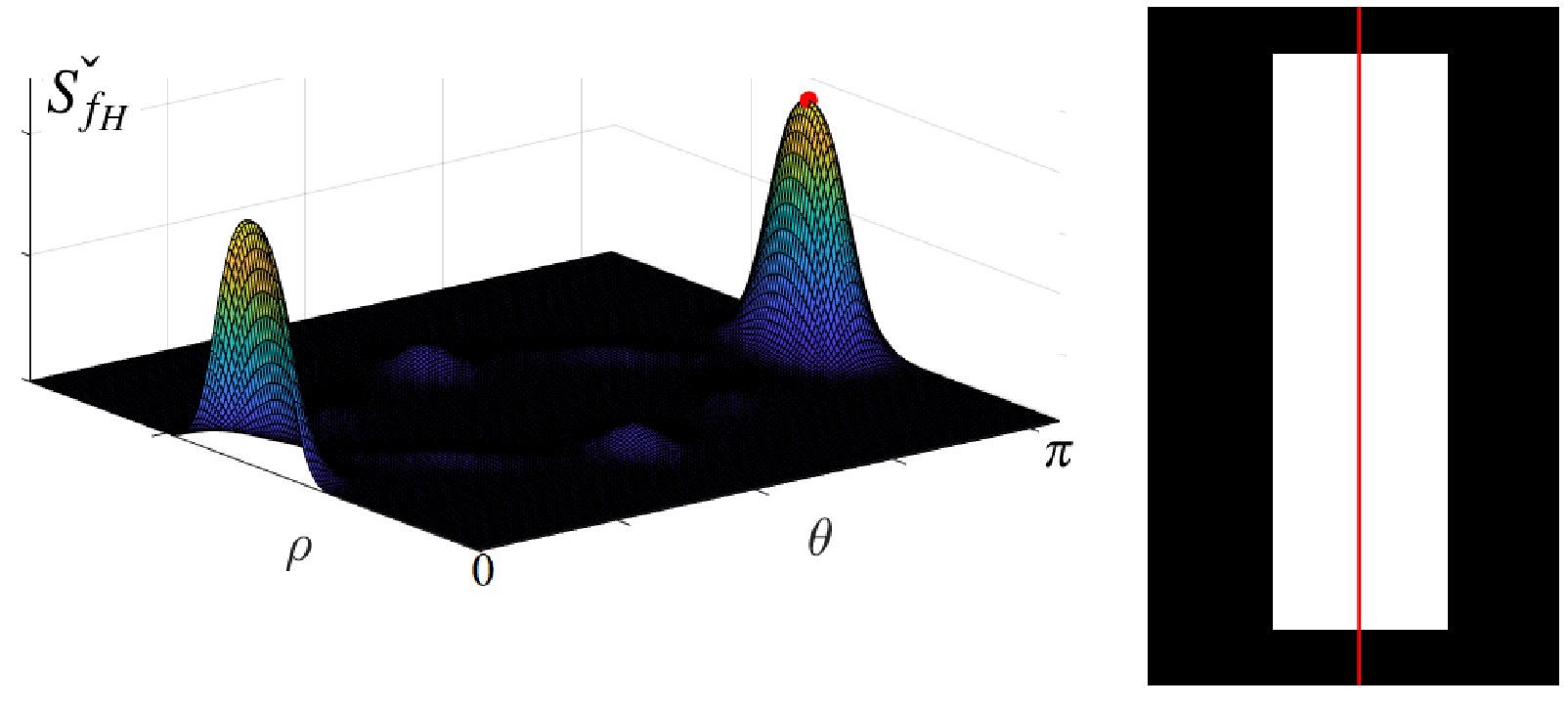}
	\caption{  Hessian guided SSRT space with $\sigma_h=3.28$ and a the detected centerline of the structure with the Hessian guided SSRT. }\label{Hess3}	
\end{figure}

The question that arises in case of exploiting the Hessian of an image is what Hessian scale should be used when no information about the linear structures width is available. It should be recalled that, the only Hessian features to be exploited here are the vectors composing $\textbf{n}^\perp$, as the SSRT computation is guided by their orientations. Let us see in the following, the effects of changing the Hessian scale, on the directions of the mentioned vectors and the consequences on the Hessian guided SSRT spaces. The linear structure shown in Fig.\ref{lambdasn}.d has as width $w=23$ pixels, so, as demonstrated in \citep{Steger}, the scale $\sigma_h$ should satisfy $\sigma_h\geq\frac{w}{\sqrt{3}} \geq13.28$, for linear structures enhancement and detection. Let us see what happens if we take scales greater and lower than this value for our guided SSRT based-detection, i.e $\sigma_h=13.28 \pm 10$. So we use $\sigma_h \in \left\lbrace 3.28, \ 23.28\right\rbrace $ and we observe the results on the Hessian guided SSRT spaces and the lines representing the detected maxima. Let us begin by $\sigma_h=3.28$. We can see in Fig.\ref{Hess1} the vectors $\textbf{n}^\perp$ for this scale. We remark that inside the structure they are pointing in the latter direction.
Concerning the maxima detection in the SSRT space, it is worth to recall that as the RT, the SSRT has aperiodicity of $\pi$, i.e. $\check{S_f}(\rho,\theta)=\check{S_f}(\rho,\theta+\pi)$, which makes us do not highlight in Fig.\ref{Hess3} the redundant maxima and do not take them into consideration. Consequently, we can notice in Fig.\ref{Hess3} that the guided SSRT space has one maximum for the structure and the detected line is located in the center of the structure as shown in the same figure. 

\begin{figure}[h]
	\centering
	\includegraphics[width=100mm]{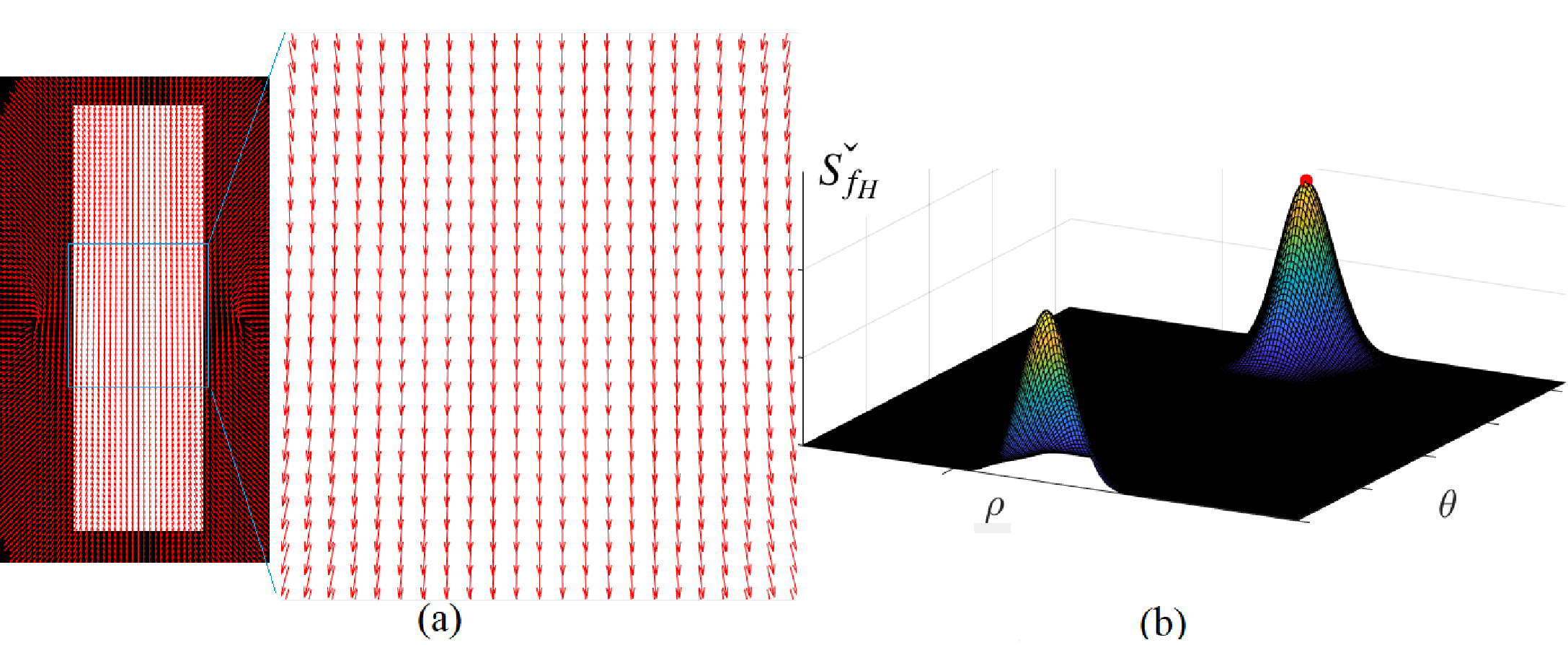}
	\caption{ (a) The field $\textbf{n}^\perp$ generated with $\sigma_h=23.28$ and a zoom on it. (b) Hessian guided SSRT space. }\label{sigma2}	
\end{figure}

\begin{figure}[h]
	\centering
	\includegraphics[width=60mm]{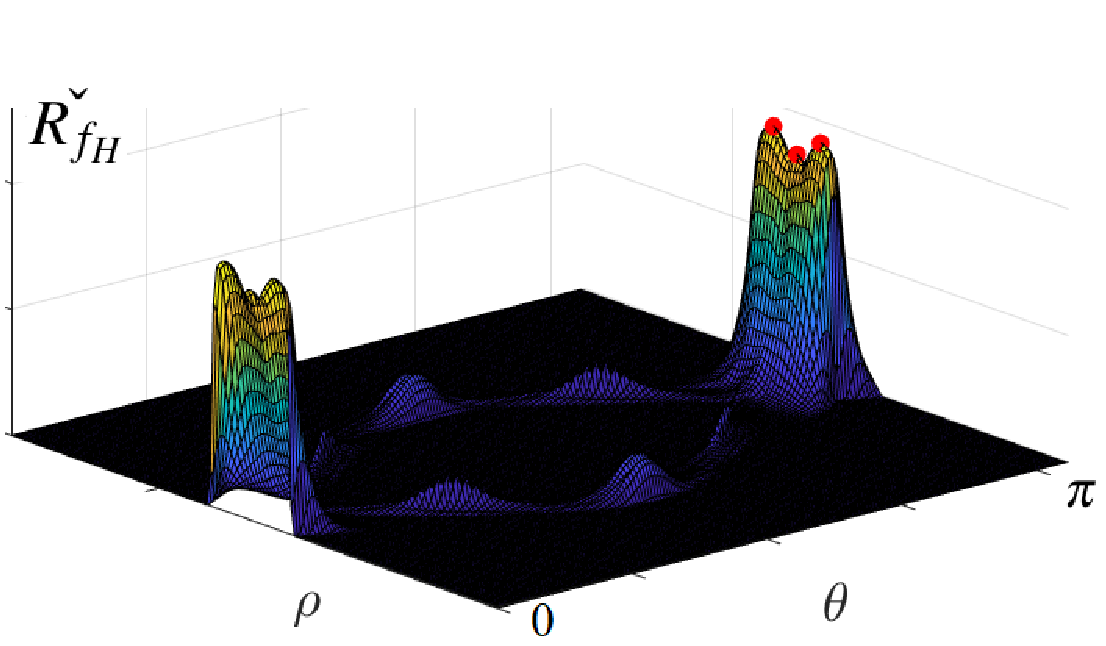}
	\caption{ Hessian guided RT space}\label{Hess2}	
\end{figure}

The result of using $\sigma_h=23.28$ is illustrated in Fig.\ref{sigma2} with the field $\textbf{n}^\perp$ in Fig.\ref{sigma2}.a and the Hessian guided SSRT in Fig.\ref{sigma2}.b. Note that, the detected centerline is the same as the one illustrated in  Fig.\ref{Hess3}. This is to say that, it is not necessary to have an exact information about the width of the linear structure to be detected, and to choose accordingly the Hessian scale. In fact, a coarse knowledge about the structure derived from the involved application, and consequently an approximate Hessian scale is sufficient to achieve detection with the proposed method, conversely to the other integral transforms when they are similarly associated to the Hessian, like the Radon transform, for example. Indeed, if we consider the Radon transform guided by the Hessian and, analogously, named it $\check{R_{f_H}}$ and computed it as $\small \check{R_{f_H}}(\rho,\theta)=\int_{\cal X} \int_{\cal Y}f_H(x,y)\delta (x\cos \theta +y\sin \theta -\rho)dxdy\normalsize$, we can illustrate the corresponding Hessian guided RT space for $\sigma_h=3.28$ in Fig.\ref{Hess2}. We can notice that the structure is represented by three maxima instead of one which confirms that RT, even associated with Hessian as done in this work, is not adapted for thick structures, as the element which allows the flexibility regarding the structure width permitting, thus, a good detection is the SSRT scale parameter $\sigma$. Additionally, the detection robustness can be enhanced by averaging the function $\Gamma(x,y)$ for two very different Hessian scales.

\subsubsection{Speeding the transform space computation up}	
The authors in \cite{Nacer} have shown that there is a relationship between RT and SSRT, expressed as follows
\begin{equation}\label{ssrt_conv}
\check{S_f}(\rho)=\check{R_f}(\rho)\circledast g_\sigma(\rho)
\end{equation}
The formulae \eqref{ssrt_conv} constitutes a straightforward and a fast way to compute the SSRT which is the convolution of the RT with the 1d kernel $g_\sigma(\rho)$ tuned by the SSRT scale space parameter $\sigma$, as  seen in Sect2. Analogously,  $ \check{S_f}_H(\rho)$, is computed via $\check{R_f}_H(\rho)$ as 
\begin{equation}\label{ssrt_convH}
\check{S_f}_H(\rho)=\check{R_f}_H(\rho)\circledast g_\sigma(\rho)
\end{equation}

\subsubsection{Practical considerations}
\begin{figure}[h]
	\centering
	\includegraphics[width=70mm]{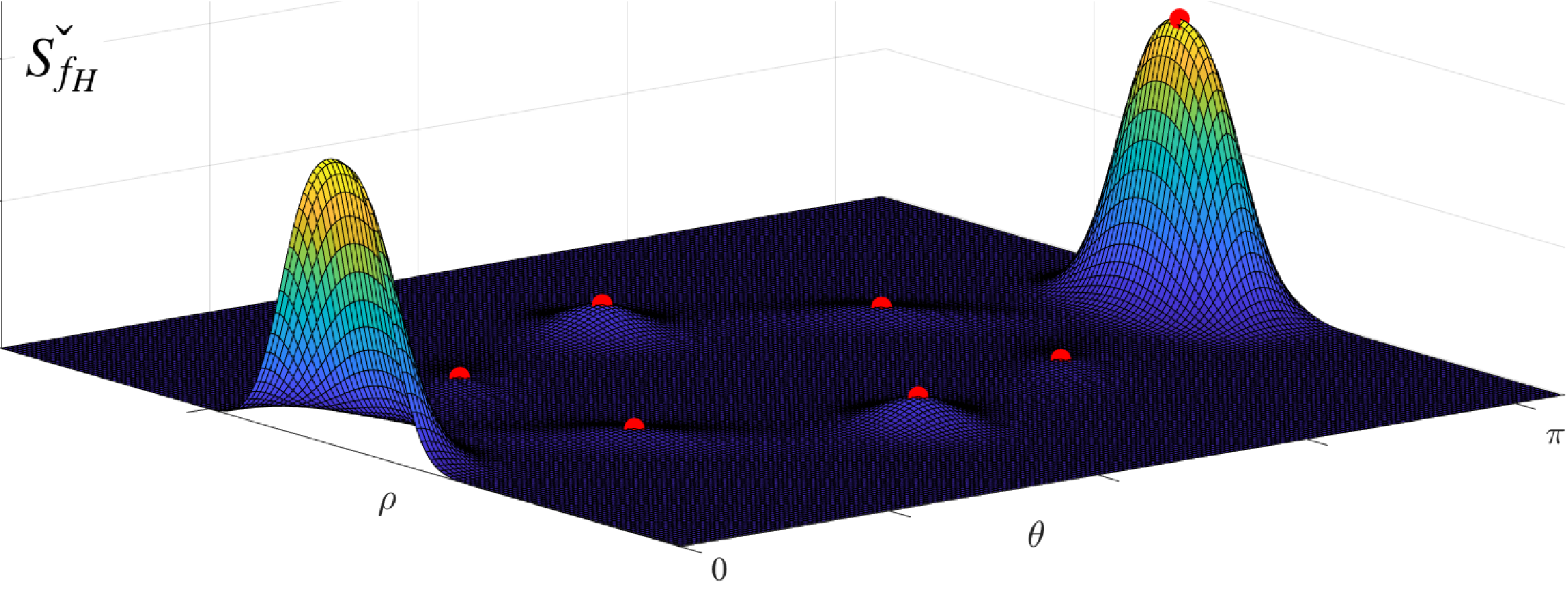}
	\caption{ Hessian guided RT space and false maxima}\label{Max}	
\end{figure}
\begin{figure}[h]
	\centering
	\includegraphics[width=60mm]{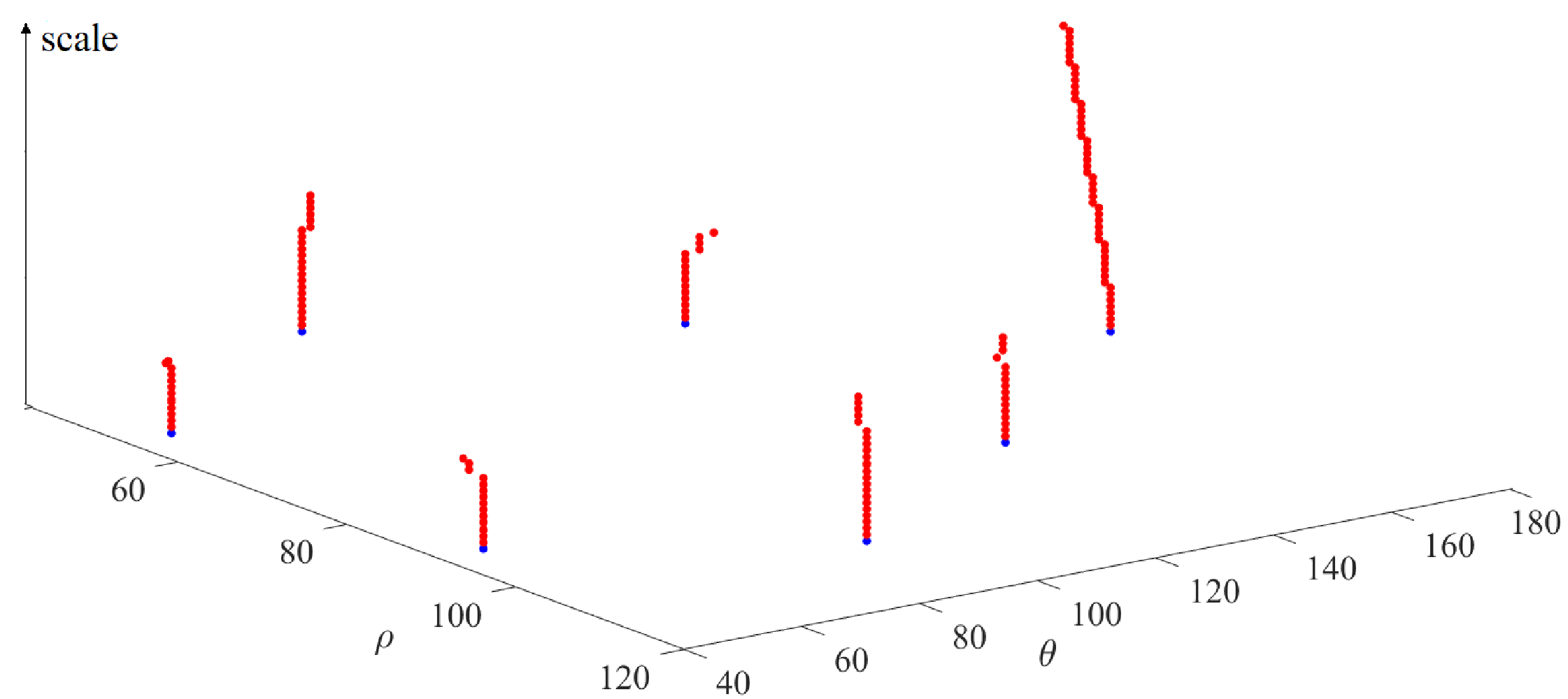}
	\caption{ Constructed 3d lines in scale space for SSRT space of Fig.\ref{Max}}\label{lines}	
\end{figure}
We can notice the presence of false maxima in Fig.\ref{Hess3}. This means that even with the application of the Hessian-guided SSRT, some irrelevant maxima remain as shown in Fig.\ref{Max}. These maxima can be removed as following. First maxima are thresholded to reject low maxima (below a threshold $TH$). Afterward, behavior in scale space of SSRT maxima above the threshold $TH$, are studied. The method applied to carry out this study is proposed in \cite{gilles} to discard non meaningful modes in histograms. By extending this study to 3d ($\theta, \rho, \check{S_{f_H}}(\theta,\rho)$) and to maxima, SSRT space is convoluted repetitively by a 3d Gaussian kernel which remove progressively local maxima and produce as well a scale space representation of all computed maxima.The number of local maxima in the scale space is decreasing with an increase of the scale step $ss$. At the end the 3d coordinates of all maxima construct 3d lines in scale space , as seen in Fig.\ref{lines}. On the basis of the corresponding line length, a maximum can be retained or rejected. This is step is considered as a SSRT space refinement stage.

\section{Experimentation and results}\label{sec4}
\begin{figure}[h]
	\centering
	\includegraphics[width=90mm]{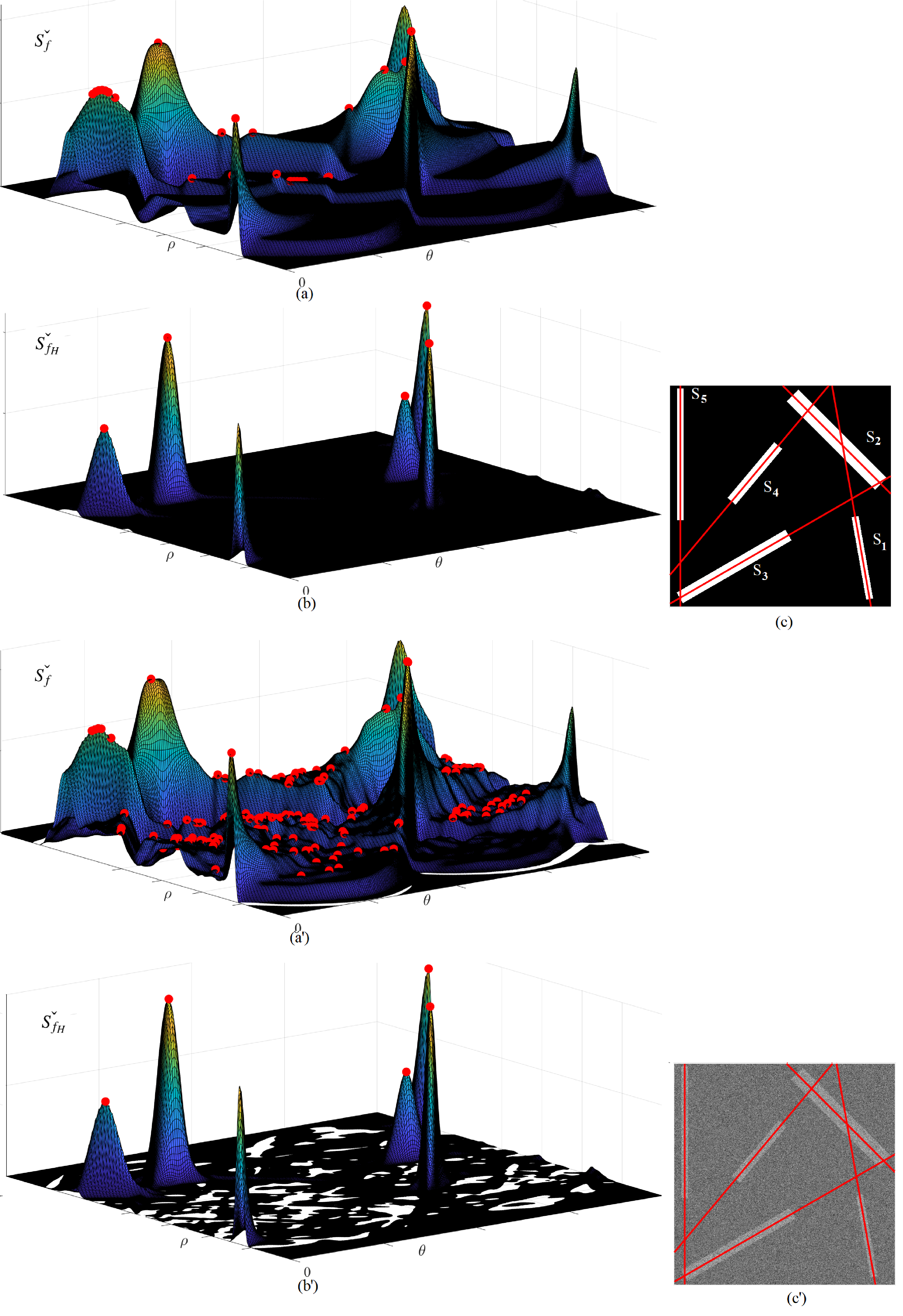}
	\caption{a: SSRT of the noise free image, b: Guided SSRT of the same image, c: Centerlines detection, a': SSRT of the noisy image, b': Guided SSRT of the same image, c': Centerlines detection}\label{res1}	
\end{figure}

To apply the proposed method, a set of synthetic and real images containing lines of one or of several pixels widths, have been used. We begin by the image of the example given in Fig.\ref{ssrtbars} and a very noisy version of it generated by adding white Gaussian noise (AWGN) with standard deviation $\zeta$ equal to 200. We have seen that applying SSRT alone even with a thresholding does not provide the desired results. In Fig.\ref{res1}, we show the result of applying the Hessian-guided SSRT method on this image and on its noisy version, in terms of guided SSRT spaces, $\check{S_{f_H}}$, and  in terms of centerlines detection. To visually appreciate the modifications undergone by the SSRT spaces $\check{S_{f}}$ when combining the Hessian and the SSRT, we depict in Fig.\ref{res1}.a and Fig.\ref{res1}.a' the SSRT spaces as well as the SSRT guided by Hessian ones, $\check{S_{f_H}}$, for both cases in Fig.\ref{res1}.b. and Fig.\ref{res1}.b'. Furthermore, the detected centerlines are provided in Fig.\ref{res1}.c and Fig.\ref{res1}.c'. The investigated images are of size $501 \times 501$, consisting of five linear structures, $S_1$, $S_2$, $S_3$, $S_4$, $S_5$ of widths equal to 15, 37,27,23 and 15 respectively. The centerlines of those structures whose equations can be written in the form $\rho_i=x\cos \theta_i+y\sin\theta_i$, are  $\left\lbrace \rho_1=362.6011,\theta_1=10^{\circ}\right\rbrace $, $\left\lbrace \rho_2=181,\theta_2=45^{\circ}\right\rbrace $, $\left\lbrace \rho_3=-428.8724 ,\theta_1=120^{\circ})\right\rbrace $, $\left\lbrace \rho_4=-276.6168,\theta_4=140^{\circ}\right\rbrace $ and $\left\lbrace \rho_5=24,\theta_5=0^{\circ}\right\rbrace $. 

To evaluate quantitatively the results, we summarize in Table.\ref*{tab_5rect}  detection accuracy in terms of absolute error values $(\Delta \rho,\Delta\theta)$ for the noisefree image, where $\Delta \rho=|\rho-\hat{\rho}|$, $\Delta \theta=|\theta-\hat{\theta}|$ and $(\Delta \rho_N,\Delta\theta_N)$ for the noisy case, where  $\hat{x}$ is the computed value of $x$. In the case of the noisy image, as the noise is generated randomly leading to slightly different images when reproducing the noise generation, the detection is then carried out forty (40) times on forty generated images and the detection error results are averaged and reported on the same table. The used Hessian scales here are $\sigma_h\in  \left\lbrace 6,16\right\rbrace$ and the SSRT scale $\sigma$ is set equal to 10. 

\begin{table}[h]
	\caption{Error values of lines detection}\label{tab_5rect}%
	\begin{tabular}{@{\extracolsep{\fill}}lllll@{}}
		\toprule
		Structure	& $\Delta\mathbf{\rho}$ & $\Delta\mathbf{\theta}$&$\Delta\mathbf{\rho}_N$ & $\Delta\mathbf{\theta}_N$ \\
		\midrule
		$S_1$ & 0.0001  & 0  &0.0004 &0  \\  
		$S_2$ & 0 & 0  & 0&0  \\ 
		$S_3$ & 0 & 0  &0.0002 &0  \\ 
		$S_4$ & 0.0001  & 0  &0.0004 &0  \\ 
		$S_5$ & 0 & 0  &0 &0  \\   
		\botrule
	\end{tabular}
	
\end{table}
In the light of the detected structures centerlines depicted in Fig.\ref{res1} that are equal to the structures number, we can notice that the centerlines seem to be drawn exactly in the center of the structure. Moreover, the results given in Table.\ref*{tab_5rect} confirming the previously visually-based assumptions, show that the proposed method provides an accurate detection of all structures in the image represented through their centerlines. Indeed, the computed detection errors are very close to zero for all structures even for those composing the very bad image quality. Moreover, we can see in Fig.\ref{res1} a., .b, a'. and .b' that the method has permitted to highlight only the linear structure in the SSRT space retaining, therefore, only the ones related to those structure and discarding simultaneously all space parts and consequently maxima that  do not represent these structure, eliminating , in addition and when computing the space, irrelevant image parts from computation which modifies considerably the SSRT space in the appropriate way as seen in Fig.\ref{res1}.b and .b' and as ascertained by numerical results. Moreover, we can notice that the used Hessian scale are not the optimal ones for the structures widths but still leads to the desired results.

\begin{figure}[h]
	\centering
	\includegraphics[width=100mm]{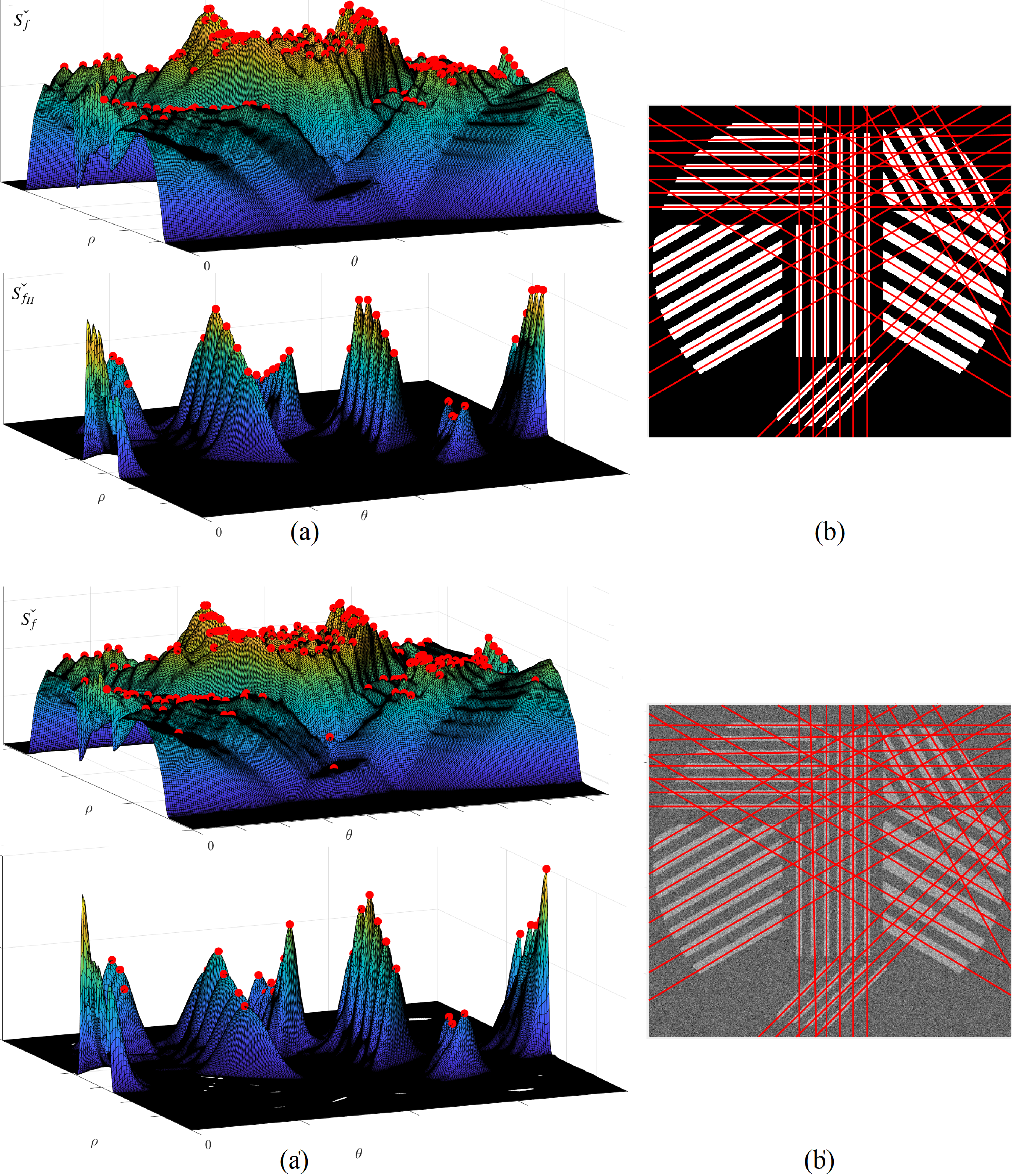}
	\caption{a: SSRT of the noise free image, b: Guided SSRT of the same image, c: Centerlines detection, a': SSRT of the noisy image, b': Guided SSRT of the same image, c': Centerlines detection }\label{res2}	
\end{figure}

The second used synthetic image is a challenging one as it composed of 35  structures of three different sizes and very close to each-other, which makes the detection a hard task when exploiting the Hessian and the SSRT. In fact, when the structures are too close to each other as it is the case for this example, each structure can interfere in the Hessian and in the SSRT computation with the adjoining structures, especially when the used $\sigma$ and $\sigma_h$ far from the optimal ones. The image structures are arranged in six different orientations and have as widths 7, 13 and 19 pixels. SSRT guided detection is performed on this image and a noisy version of it, where the noise is a white Gaussian noise of standard deviation $\zeta=100$. Results of such application are shown in Fig.\ref{res2} and in Table.\ref*{tab_34}, in terms of root mean square error (RMSE), noted $\sum_\Delta$ and computed as  $\sum_{\Delta\theta}=\sqrt {\sum_{i=1}^{i=N}(\theta_i-\hat{\theta}_i)^2/N}$ and  $\sum_{\Delta\rho}=\sqrt {\sum_{i=1}^{i=N}(\rho_i-\hat{\rho}_i)^2/N}$ where $N$ is the number of the structures. In consideration of Fig.\ref{res2} and Table.\ref*{tab_34}, the noise free image detection seems to be accurate despite the structures promiscuity. Indeed, as illustrated in  Fig.\ref{res2}.c, the detected centerline seem to be well located. This fact is ascertained by the detection errors that are very low and approximate the zero value for both $\theta$ and $\rho$, as shown in the mentioned table. Moreover, we can see how the guided SSRT looks in Fig.\ref{res2}.b comparatively to the SSRT in Fig.\ref{res2}.a which shows that joining the structures orientation by means of the Hessian the way we proceed, has operated a big modification to the SSRT space by emphasizing the linear structures while removing all unwanted interference's. Nevertheless, regarding the noisy image, the detection errors have slightly increased especially for the parameter $\rho$, which means that the detected centerlines have been subjected to displacement. However, the detection can considered as acceptable for this last case since the image is of a poor quality. In conclusion of the previously tests we can conclude that having structures too close to each other can alter the detection for very poor quality images, even if this detection can be considered as acceptable. To end tests for synthetic images we can note that one of the structures has not been detected. It is the smallest one. This is due to the fact that the linear part of it is too small to be assimilated to a linear structure as it has a very low maximum in the guided SSRT space and has been, consequently, deleted in the refinement stage. To finish for this experiments, the used Hessian scales here are $\sigma_h\in  \left\lbrace 3,8\right\rbrace$ and the SSRT scale $\sigma$ is set equal to 5.

\begin{table}[h]
	\caption{RMSE for line detection based on Guided SSRT}\label{tab_34}%
	\begin{tabular}{@{\extracolsep{\fill}}lll@{}}
		\toprule
		Image	& $\sum_{\Delta\mathbf{\rho}}$  & $\sum_{\Delta\mathbf{\theta}}$ \\
		\midrule
		Noise free image 	& 0.7678  & 0.7071\\
		Noisy image  &2.8150 &0.811 \\  	
		\botrule
	\end{tabular}
	
\end{table}

\begin{figure}[h]
	\centering
	\includegraphics[width=110mm]{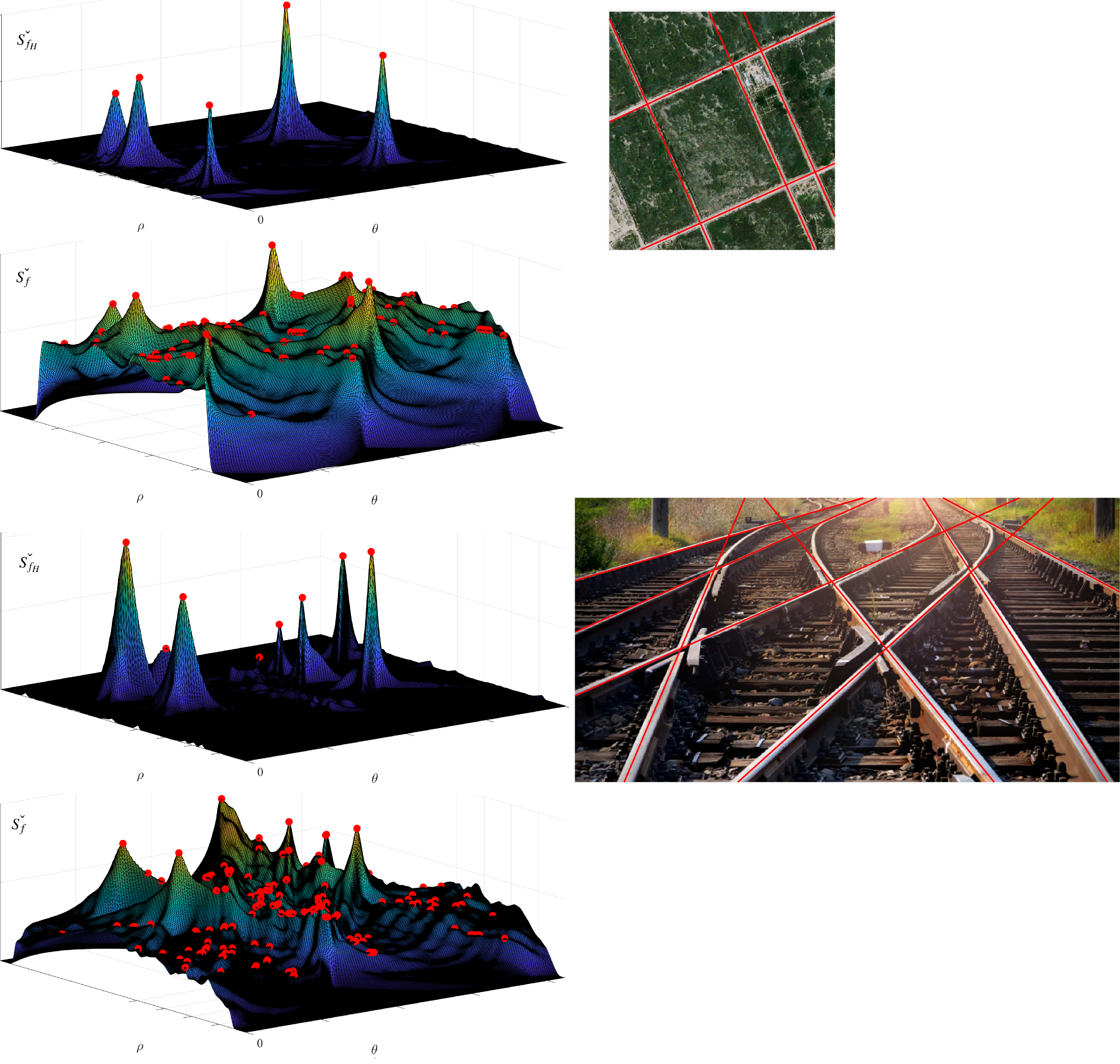}
	\caption{From left to right:  Guided SSRT and SSRT spaces and the corresponding paths detection in archaeological site and railroad detection}\label{res4}	
\end{figure}

\begin{figure}[h]
	\centering
	\includegraphics[width=100mm]{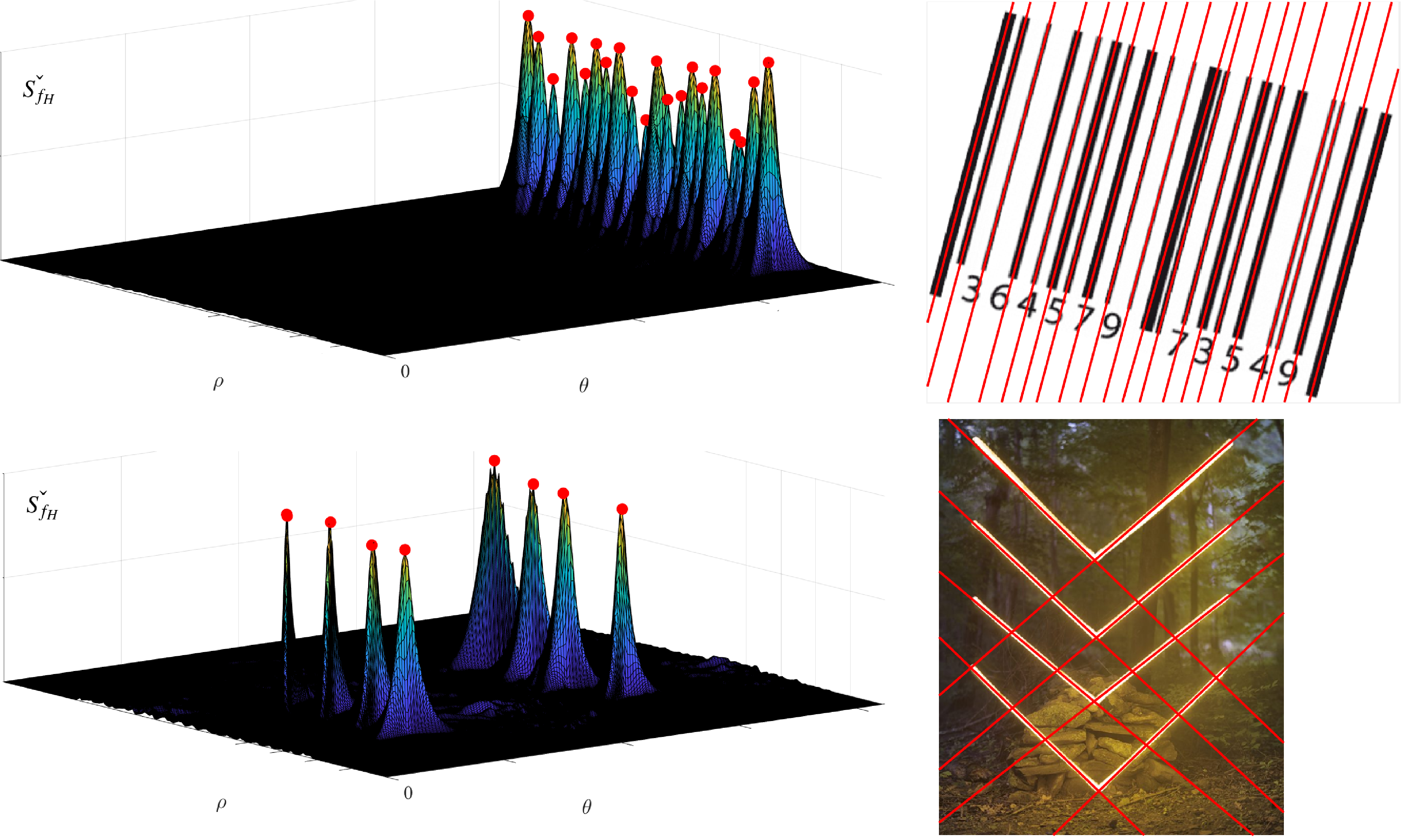}
	\caption{From left to right:  Guided SSRT spaces and the corresponding bar-code and neon detection }\label{res3}	
\end{figure}

\begin{figure}[h]
	\centering
	\includegraphics[width=100mm]{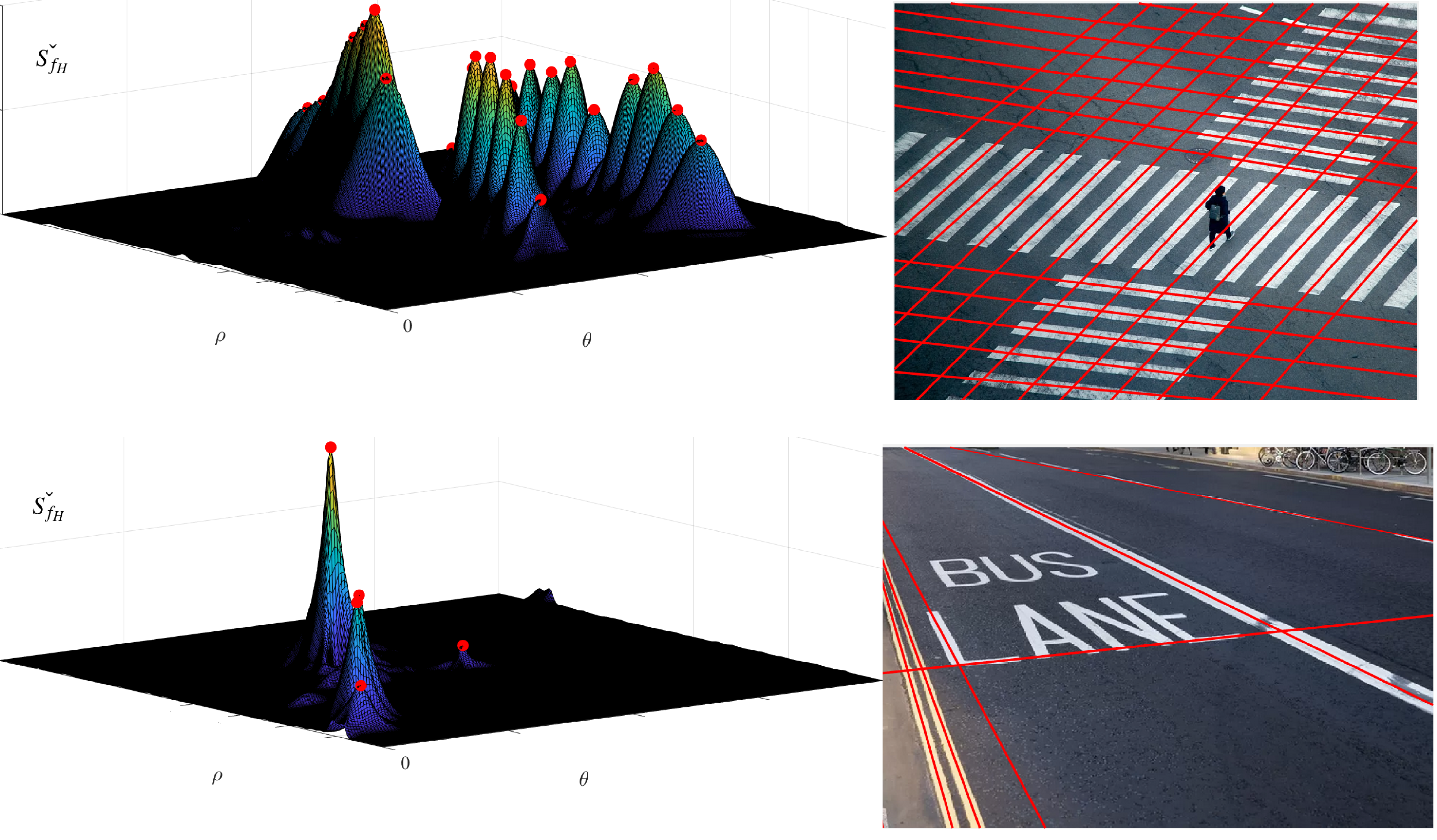}
	\caption{From left to right:  Guided SSRT spaces and the corresponding crosswalks and lane detections}\label{res5}	
\end{figure}

Next experiments have been done on several real images including bar-code,paths in an archaeological site, railroad, crosswalks and lanes given in\ref{res4},  Fig.\ref{res3} .and .\ref{res5}, where we can see the guided SSRT in each case along the structures centerlines detection. In \ref{res4} $\check{S_f}$ spaces are depicted with $\check{S_{f_H}}$ ones to see how a complex background is reflected in the SSRT space for real images and the amount of irrelevant maxima contained in the SSRT space. We can notice, also, that for certain structures the detected lines are not in the center of those structures. This is due to the fact that the color is not uniform inside the structure as we can see for the crosswalks image, for example, or for the railway one. Moreover, the structures edges are irregular and not smooth, which influences the SSRT and the Hessian computation. Concerning the detection parameters, the used Hessian scales here are $\sigma_h\in  \left\lbrace 3,8\right\rbrace$ and the SSRT scale $\sigma$ is set equal to 5 except for the image of crosswalks as the structures are very thick, where $\sigma$ is set to 10. For Color images, all computations are performed on gray scale version of them.

\section{Conclusion}\label{sec13}

In this paper, we have proposed to combine the SSRT and the Hessian of the image to tackle, in an elegant manner, the issue of linear structures detection in images in presence of noise and complex background. The method modifies, via the Hessian, the SSRT computation so that its failing in the detection when facing complex images is alleviated. Indeed, this is done by making the SSRT projections effective only for features having the same orientation as the projection direction. In fact, failing in SSRT detection arise when linear structures are too close to each others. Moreover, when the investigated structures are drown in a complex background, their corresponding maxima in the SSRT space are also drown in a huge amount of other false maxima. To overcome these drawbacks, all image features, including linear structures that do not have the actual orientation of the SSRT projection are considered irrelevant and therefore attenuated. The others are enhanced to produce maxima that are easy to detect in the modified SSRT space. Experimentation carried out on both synthetic and real images confirms the effectiveness of the proposed method. In fact, besides correct location of the structures centerlines, it has proven its robustness against noise and complex background.
\subsection*{Availability of data and materials}
The datasets used and analyzed during the current study are available from the corresponding author upon reasonable request 
\bibliography{ABG_refs}
\end{document}